\documentclass[10pt,twocolumn,letterpaper]{article}

\usepackage{cvpr}
\usepackage{times}
\usepackage{epsfig}
\usepackage{graphicx}
\usepackage{amsmath}
\usepackage{amssymb}
\usepackage{cite}

\usepackage[breaklinks=true,bookmarks=false]{hyperref}

\cvprfinalcopy 

\def\cvprPaperID{****} 

\newcommand*\rot{\rotatebox{60}}

\begin{document}
	\title{Asian Stamps Identification and Classification System}
	
	\author{Behzad Mahaseni\\
		Multimedia University\\
		{\tt\small b.mahaseni@gmail.com}
		\and
		Nabhan D. Salih\\
		Multimedia University\\
		{\tt\small nbhan@mmu.edu.my}
	}
	
	\maketitle
	
	\begin{abstract}
		In this paper, we address the problem of stamp recognition. The goal is to classify a given stamp to a certain country and also identify the year it is published. We propose a new approach for stamp recognition based on describing a given stamp image using color information and texture information. For color information we use color histogram for the entire image and for texture we use two features. SIFT which is based on local feature descriptors and HOG which is a dens texture descriptor. As a result on total we have three different types of features. Our initial evaluation shows that give these information we are able to classify the images with a reasonable accuracy.

	\end{abstract}
\section{Introduction}\label{sec:introduction}
Intelligent stamp recognition system is an essential part of the new modern post offices. Automatic mail grouping is one of the many applications of stamp recognition in the postal services. In addition museums and antique collectors can use intelligent stamp recognition software to help them categorize stamps to pre-known classes. Motivated by the above we believe it is important to study the current approaches for stamp recognition. 

The proposed system should be capable of recognizing different stamps from different countries. At the basic level, it should be able to recognize the country of origin for a given stamp. In the advanced mode, it should be able to recognize the year which the stamp is published.  To do so it should use some basic features and characteristics from the stamp image (e.g. color distribution, texture information) to describe a given image. In this project, we aim to develop a prototype of such a system.
To make it more feasible, we limit the focus of the project to the stamps published from 2010 – 2015 from five Asian countries. As a result one important part of this project is to study the current stamp dataset (if there is any available) or to create a new dataset for this purpose. 

Stamp recognition is a challenging problem. This is mainly due to the high intra-class similarity of different stamp classes and low-resolution stamp images. In addition, the stamp might not be completely visible due to physical destruction of the paper. Since the standard, rule-based systems are not capable of handling such a challenging problem, machine learning techniques seem to be a more reasonable choice. It is important to note that this problem is a fine-grained classification problem because the visual properties of the stamp classes are so similar. 

Unlike traditional AI, in machine learning the goal is to learn a model based on a set of training examples, where each  is one instance. The model should be learned such that it generalizes well to the examples which have not been observed during training phase. This is called generalization. 

To summarize the objectives of this paper is as follows:
\begin{itemize}
	\item Study and analyze different feature extraction techniques.
	\item Study different classification models
	\item Evaluate different combinations of the features and classifiers
\end{itemize}

In the rest of this report, we provide a brief review of the currently available stamp recognition systems in sec.~\ref{sec:prior_work}. We then review the theoretical aspects the classification framework for which our system is built on in sec.~\ref{sec:model}. Later in sec.~\ref{sec:features} and sec.~\ref{sec:classification-models} we introduce details of feature extraction and classification models. In sec.~\ref{sec:dataset} we provide details of the new dataset we create from online stamp images.  Sec.~\ref{‎sec:experiments} shows different test and evaluation results on our dataset.  

\section{Related work} \label{sec:prior_work}

In this section we review the current stamp recognition software and we as well review the basic machine learning concepts. Review of the basic machine learning concepts is important mainly because later we use the notation introduced in this section.

\subsection{Stamp Recognition Literature }
	There is little prior work in research community on stamp recognition. Authors in (Li, 2011) \cite{li2011postage} use color information to identify color grouping of new stamps using template matching. In our work we want to classify the stamps based on their country and year not their color group. (Forczmnski, 2010) \cite{forczmanski2010stamp} proposes a three step algorithm for stamp detection in scanned images. Later they use Fast Fourier Transform for feature representation and used the nearest neighbor matching to find the closest stamp in the dataset to verify the resulting detection is a stamp. Also the last stage of their approach classifies with respect to current classes their main goal is to detect the stamp.
	
\subsection{Stamp Recognition Softwares}

“SRS” (Stamp Recognition) is one of the well-known commercial stamp recognition software. Based on their product introduction, this software allows you to quickly scan your stamp and it will then tags your stamp intelligently and stores it based on the assigned tags. Similar to SRS we provide stamp recognition functionality. Our goal is to provide similar functionality in an academic project. Similar to SRS we want to identify properties of the stamps (e.g. country, year, …). Unlike them, we do not have access to commercial stamp datasets and we need to provide a new dataset based on a new collection of stamps. In addition, we also want to analyze different algorithms and report the accuracy on each.

"Lignup" is a commercial stamp identification software. Similar to SRS it provides stamp search and verification functionality. It also provides a huge collection of stamps. Similar to this application we also provide a GUI for the users to upload their scanned images. 

\section{Model}\label{sec:model}

As stated in section ‎0 “Stamp Recognition” as the name suggest is a recognition problem which has been addressed extensively in AI and Machine Learning community. Our goal is to provide a software which lets the uses automatically tag a given image with the country label and the year of publication. The rest of this section provides an adequate summary of the classification approaches in general and the most common approaches in image processing and computer vision. We first review the most recent and successful image descriptors in computer vision and image processing community and provide example successful applications of these algorithms. Later we review the two important classification models which have been successfully applied in computer vision community. 

	We also introduce our new collected dataset and provide some examples from this dataset at the end of this section.
	
\subsection{Image Features}	\label{sec:features}
Given an image it is easy for humans to understand the content in an image. For example given an image it is easy fro human to identify faces in the image. Unlike humans, computers do not automatically understand the content in an image. From a computers program perspective, an image is an ordered set of pixel values. For a gray-scale image, is a two dimensional array where each element stores the pixel intensity. For a color  image, the image is actually three two-dimensional arrays where each array stores the intensity values for one of the red, green or blue channel. 
	
\begin{figure}
	\centering
	\includegraphics[width=0.5\textwidth]{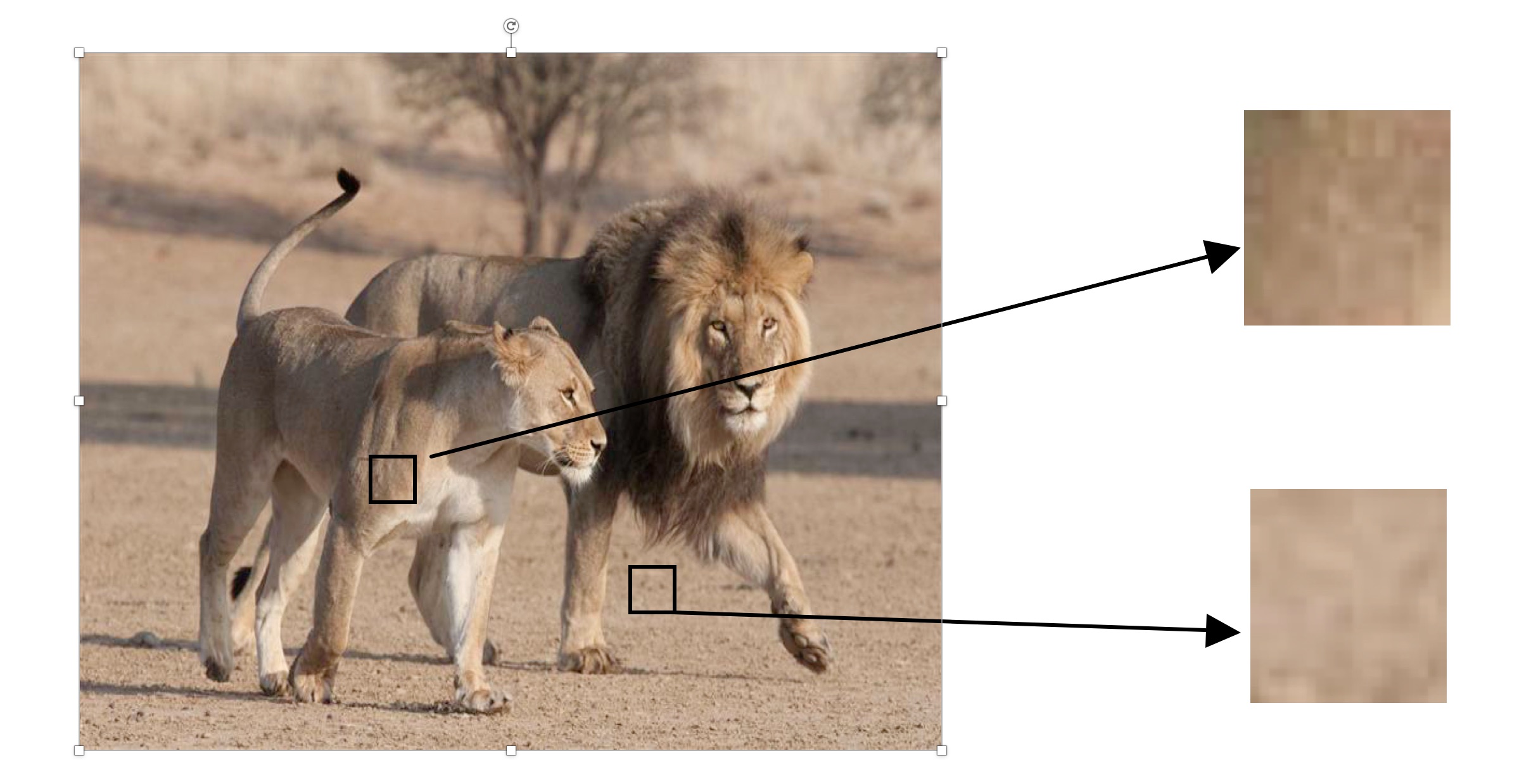}
	\caption{Example of two different patches. (a) Original lion image. (b) An image patch from the lion (c) an image patch from the soil.}
	\label{fig:fig1}
\end{figure}

Unfortunately it has been shown that single pixel intensities are not informative for global reasoning about an image. Consider the following image patch. By only observing this image patch it is hard to identify what is the image. This is mainly due to the fact that local appearances of small regions are highly stochastic and the very fact that neighboring patches have high correlation. In addition, not all of the information in an image is useful for recognition. Most of the pixel values are not informative and they do not contain helpful information for the recognition task. This has inspired researchers to introduce more global image descriptors which summarize the entire image into much smaller feature descriptors. Fig.~\ref{fig:fig1} shows this problem.

The next three subsections review the three most useful image descriptors. 

\subsubsection{Color Histogram}

Based on the Wikipedia "A histogram is a graphical representation of the distribution of numerical data". It is a mechanism to provide the empirical distribution of a stochastic event. The simplest way to represent an empirical distribution is to count the number of occurrence of each possible value. 

It is always convenient to work with the normalized values of the data. In the case of histograms the normalized values are the frequency of occurrence. Also if the values are continues or the number of discrete values are large, the standard technique is to define certain number of bins. 

The color histogram is defined as the frequency of occurrence of pixel intensity values, i.e. considering the range (0, 255) possible values for a single channel image; one can provide a histogram of the frequency of observing a certain value in the entire image. It has been shown that the color histogram is a very useful descriptor of the entire image and in certain application it can provide an accurate descriptor. 

\subsubsection{SIFT (DIASY)}
Studies in cognitive science have shown that certain points in an image are considered more informative for humans. In the image processing and computer vision terminology these are referred to as "interest points". Corners and edges have been used as interest points in computer vision. In addition to identifying the interest points, one should also describe it. Several approaches have been proposed in the past years for feature detection and description. Among these approaches SIFT is the mostly used one. 

\begin{figure}
	\centering
	\caption{Visual representation of SIFT. \cite{Lowe2004}}
	\includegraphics[width=0.5\textwidth]{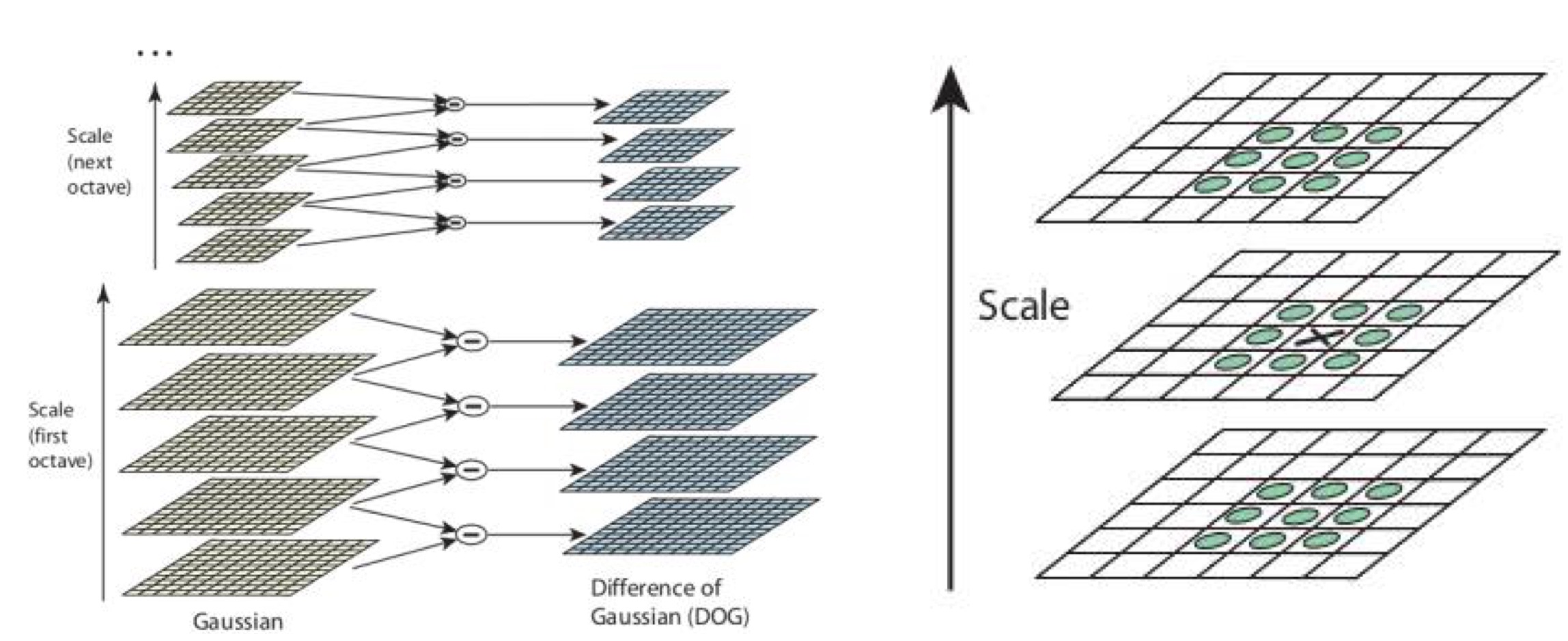}
\label{fig:fig2}
\end{figure}

Scale-invariant feature transform (or SIFT) (Lowe, 2004) \cite{Lowe2004} is an algorithm in computer vision to identify and represent local features in images. The main difference between SIFT and other interest point detectors is that SIFT uses a pyramid of different image sizes and use a blob detector to identify potential interest points in each scale. Fig.~\ref{fig:fig2} shows the image pyramid and the approximate blob detector used in SIFT.

The interest-point descriptor is a 16x16 neighborhood around the interest-point. It is split into 16 sub-blocks of 4x4 size. For each sub-block, 8 bin orientation histogram is created. So a total of 128 bin values are available. This is used as a descriptor for each feature point. Figure 4 shows the SIFT descriptor.

\begin{figure}
	\centering
	\includegraphics[width=0.5\textwidth]{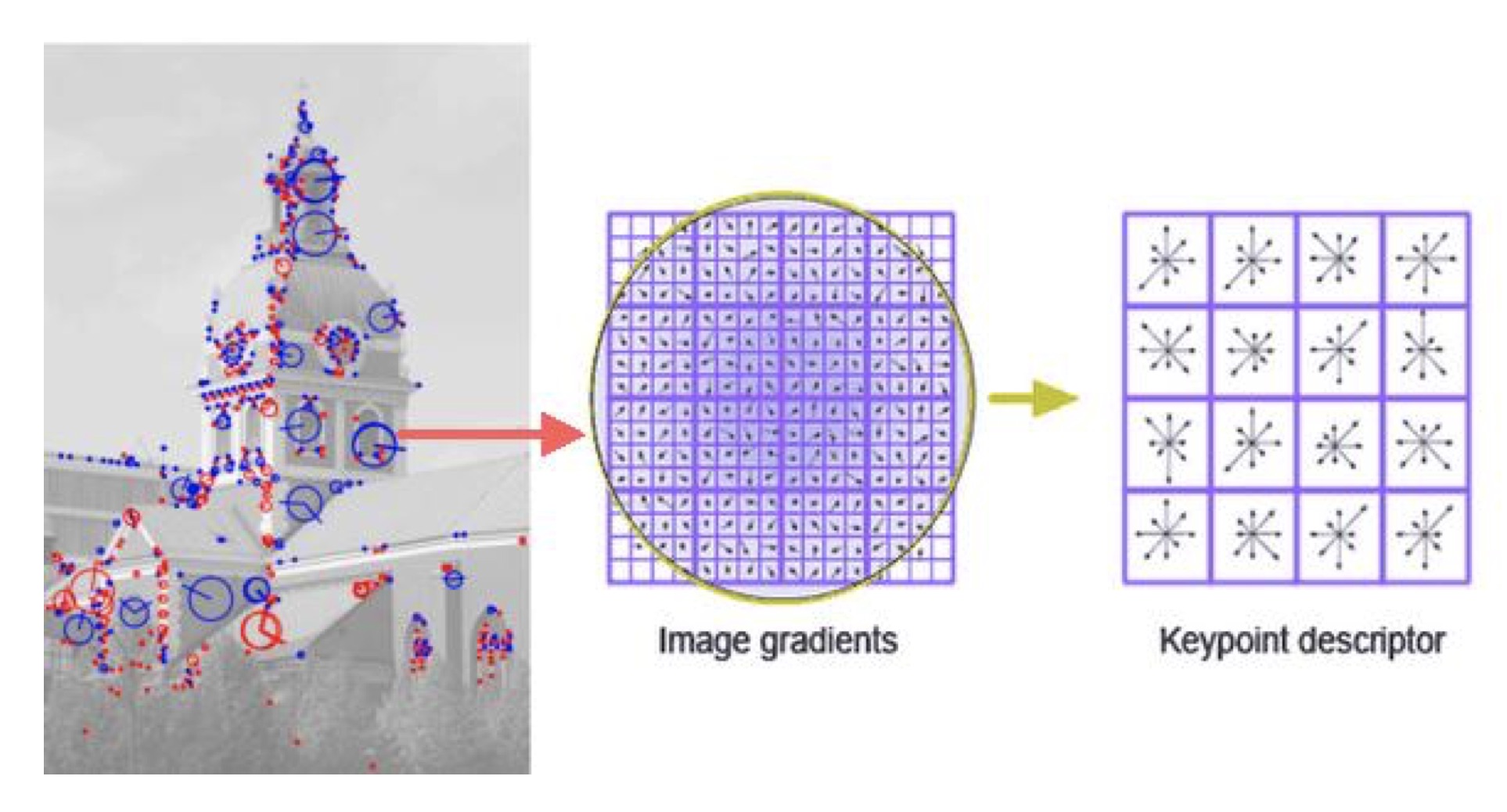}
	\caption{Visual representation of SIFT features and the descriptor . \cite{Lowe2004}}
	\label{fig:fig3}
\end{figure}
\subsubsection{Histogram of Oriented Gradients}
	Unlike SIFT which first find the interest points and then use a descriptor to summarize each interest point, Histogram of Oriented Gradients (HOG) (Triggs, 2005)\cite{1467360} is a texture based feature descriptor which provides a single description of the entire image. In other words it generates a dense descriptor of the entire image. This property makes it unique for object recognition purposes. The most successful application of HOG is in the problem of pedestrian (person) detection. 
	
Intuitively it seeks to obtain the shape of the pattern in the area by capturing information about gradients or the edge orientations. It does so by splitting the image into small (e.g. 8x8 pixels) cells and blocks of 4x4 cells. Each cell has a predefined number of gradient orientation bins. Each pixel in the cell votes for a gradient orientation bin with a vote proportional to the gradient magnitude at that pixel. Figure shows a simple illustration. 

\begin{figure}
	\centering
	\includegraphics[width=0.5\textwidth]{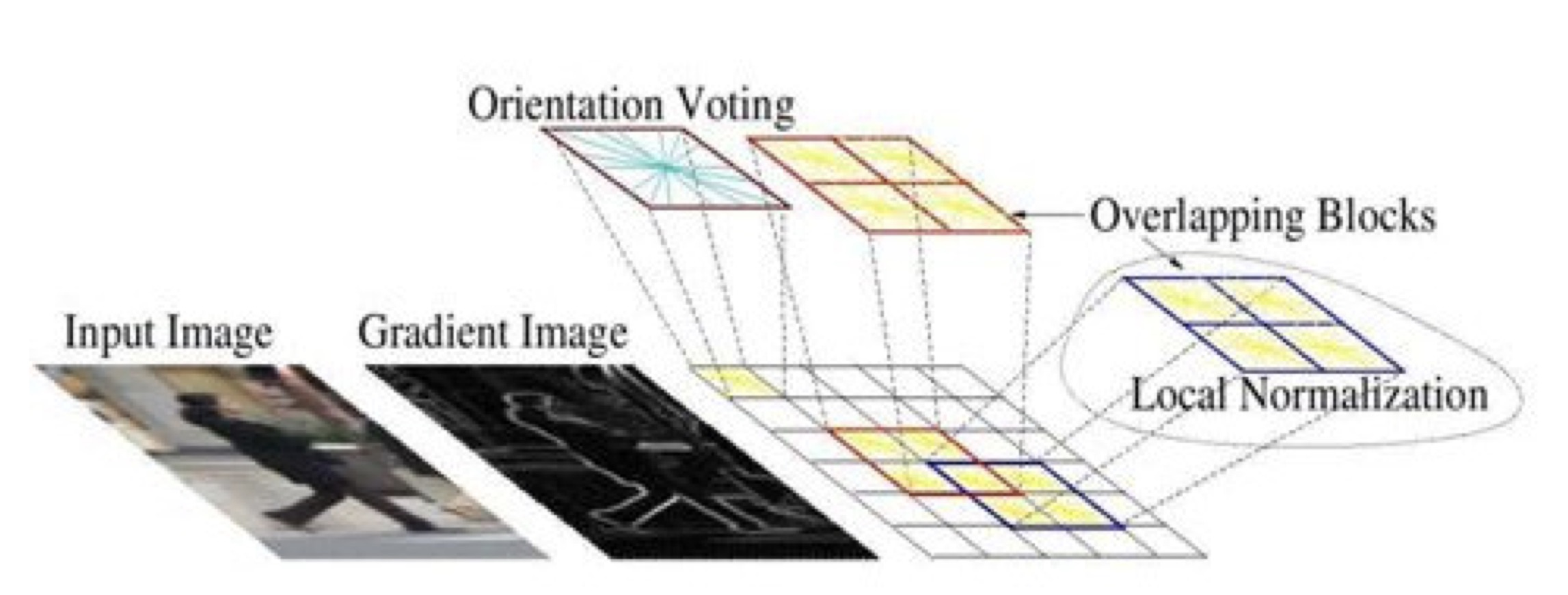}
	\caption{Visual representation of Histogram of Oriented Gradients . \cite{1467360}}
	\label{fig:fig4}
\end{figure}

\begin{figure}

	\centering
	\includegraphics[width=0.5\textwidth]{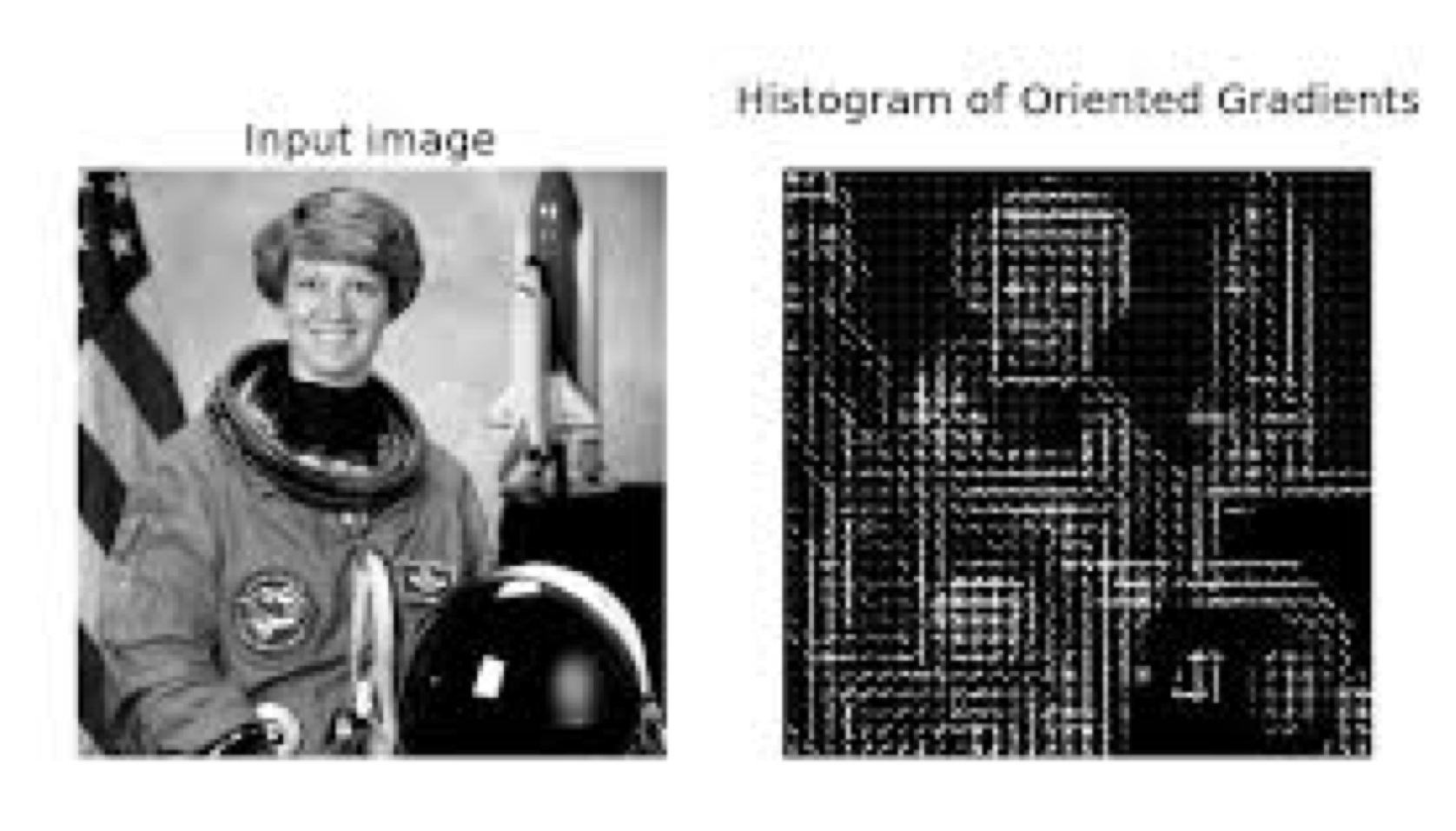}
	\caption{Example image and its corresponding HOG image. \cite{1467360}}
	\label{fig:fig5}
\end{figure}

Authors of the paper also visualized the values, fig.~\ref{fig:fig4}, and the votes of each pixel and use this visualization to show how HOG features provide texture information. fig.~\ref{fig:fig5} shows a sample visualization of HOG for an image with a person.

\subsection{Classification Models}	\label{sec:classification-models}

Given the specifications of the problem we are addressing, "Stamp Recognition" is categorized as a supervised classification problem. The main reason is that the predictions are discrete values such as countries and years. The values are not fractional numbers. In addition in training phase, for each sample we have access to the annotated country and year.

As a supervised machine learning approach, classification requires the learning algorithm has access to labeled input and output pairs. More specifically given a dataset, , the goal is to learn a function , which maps every input to a single output, i.e. . If the  values are discrete the problem is considered as a classification problem. If the values are rational numbers it is considered a regression problem. A famous classification problem is the "spam-detector". Given an email the model assigns a "spam" or "non-spam" label to it. The most famous regression problem is the least square problem. 

In this paper we have applied two different classification approaches as follows:

\subsubsection{Support Vector Machines}
\begin{figure}

	\centering
	\includegraphics[width=0.5\textwidth]{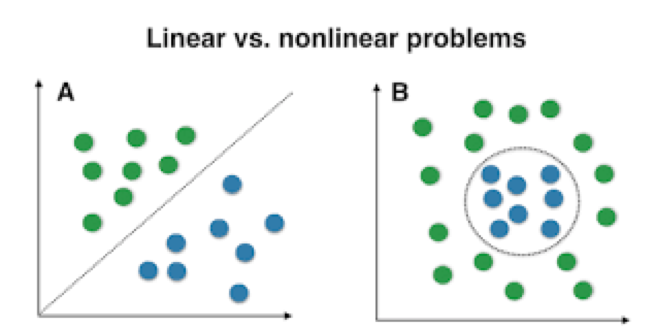}
	\caption{Visualization of linearly separable vs. not linearly separable problem}
	\label{fig:fig6}
\end{figure}

Support vector machine (SVM) (Cortes, 1995) \cite{Cortes1995} is one of the most successful classifiers in the history of discriminative models. The basic idea in linear SVM is that if you have a linearly separable data, i.e. you can draw a line between two categories and separate them. Fig.~\ref{fig:fig6} shows the different between a linearly separable and not linearly separable data.

\begin{figure}

	\centering
	\includegraphics[width=0.5\textwidth]{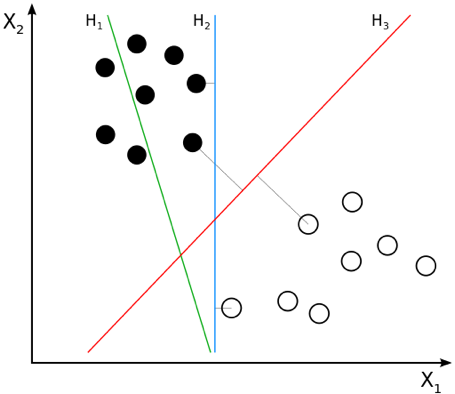}
	\caption{Difference between different lines for separating linearly separable data.}
	\label{fig:fig7}
\end{figure}

For a given linearly separable dataset, there exist several different lines which can separate the classes. Different classifiers differ in the way they define the line. SVM chooses the most intuitive line, the line which has the largest margin from both classes. This is shown in fig.~\ref{fig:fig7}. In the figure line does not separate the two classes, separates the two classes with a very small margin which can be problematic for generalization and separates the two with a more secure margin. 

The basic idea behind SVM is to form a linear function of the form $f(x) = w^\top x$ where s the model parameters vector. From the geometric perspective $\frac{1}{\|w\|}$ can be interpreted as the margin, where ${\|w\|}$ is the  $l2$-norm of the  vector. So the goal in training SVM is to find the parameters $w$ such that $\frac{1}{\|w\|}$ is maximized and all classes are classified correctly.

\subsubsection{Logistic Regression}

\begin{figure}
	
	\centering
	\includegraphics[width=0.5\textwidth]{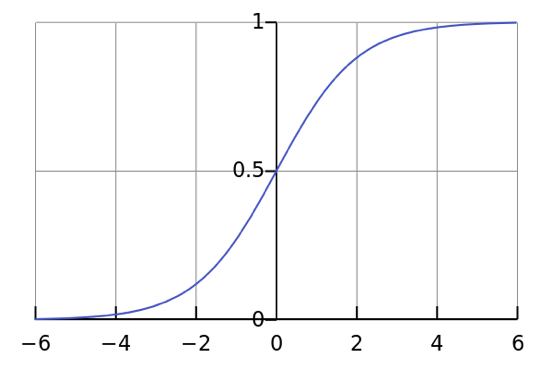}
	\caption{Sigmoid function}
	\label{fig:fig8}
\end{figure}

Logistic regression is one of the oldest probabilistic classifiers. Unlike the SVM, which only provides unnormalized score associated with each class, the logistic regression provides a valid probability distribution over the class labels using a non-linear sigmoid function. As it is shown in fig.~\ref{fig:fig8} the domain of sigmoid function is all real numbers and the output is a number between 0 and 1. More importantly the area under the integral sums to 1 which makes it a proper distribution. 

Similar to the SVM, logistic regression also forms a linear function of the form $f(x) = w^\top x$, but unlike SVM it uses the sigmoid function defined above to normalize it. This results in $f(x) = \sigma(w^\top x)$ .

\section{Stamp Dataset}\label{sec:dataset}

We could not find a benchmark dataset available for research for stamp classification. Also we wanted to only classify stamps from four Asian countries. To do so we use the stamps provided in colnect website \footnote{http://colnect.com/en/stamps/countries}. This website provides access to stamps from more than 200 countries for more than 50 years. We spend 20 hours to access the stamps for China, Japan, Malaysia, Singapore, and South Korea. We collect stamps from 2011 – 2015. For each year on average we have 70 different stamps and totally we collect 1520 stamps. 

\begin{figure*}
	\centering
	\includegraphics[width=\textwidth]{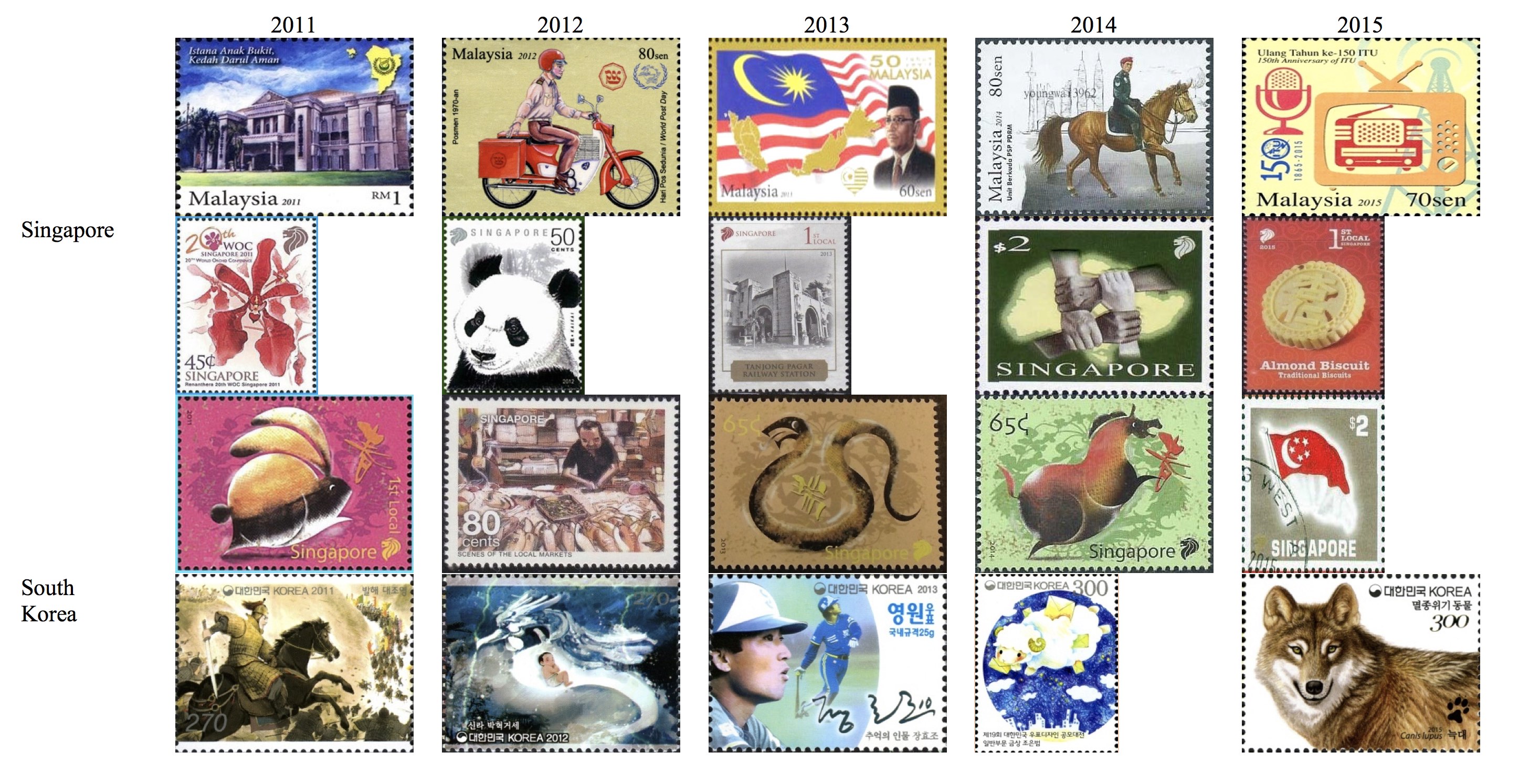}
	\includegraphics[width=\textwidth]{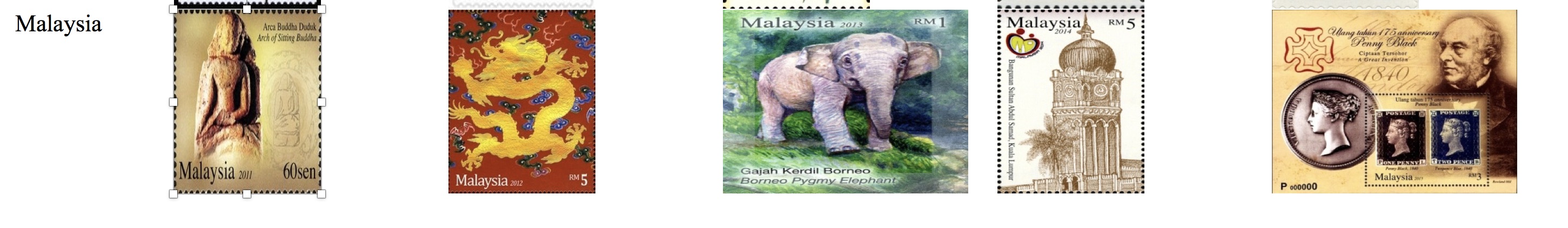}
	\includegraphics[width=\textwidth]{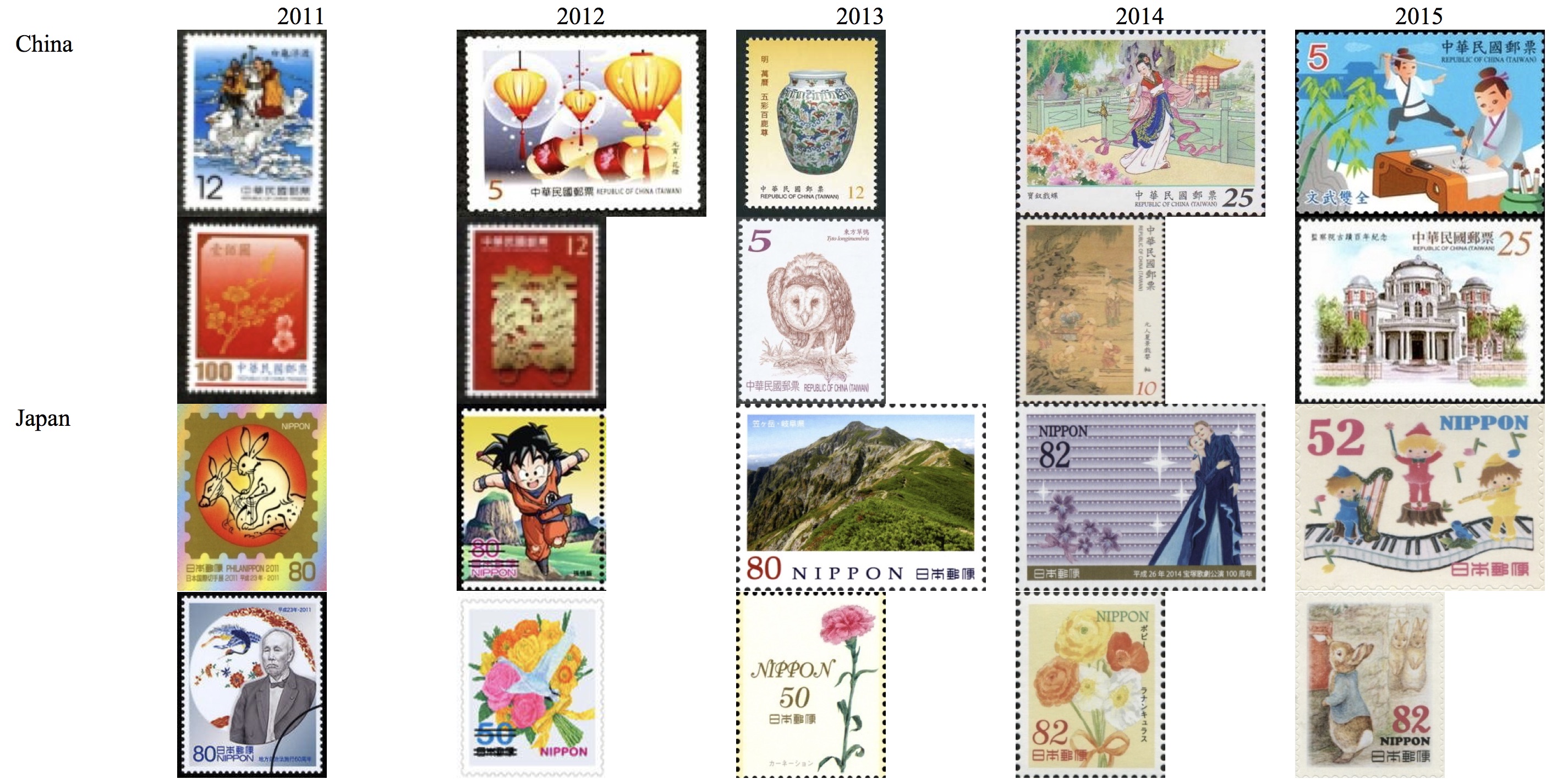}
	\newline
	\caption{Sample stamp images from five countries for five years. The images are downloaded from colnect}
	\label{fig:fig9}
\end{figure*}

Fig.~\ref{fig:fig9} shows sample stamps from these countries. As it is shown stamp images from different country look so similar. This means that we have a high inter-class similarity and the classification problem is a fine-grained recognition problem. 

\section{Experiments} \label{sec:experiments}
In this section we provide more details of the implementation as well as the obtained results through our model. 

\subsection{Implementation}

The implementation of this project is done in “Python”. The main reason for choosing python is Python language provides a rich set of libraries for computer vision and machine learning. We used “Python Image Library” (PIL) \footnote{http://www.pythonware.com/products/pil/} and “skimage” library for loading images and extracting features. We used the “sklearn” library for machine learning algorithms. 
For GUI implementation we used platform independent “Tkinter” library from python. We used “eclipse” development platform which provides rich functionality for python development through “pydev”. 
We refer the reader to Appendix A, for details of the implementation.

\subsection{Results}
In this section we present two important results we obtained from evaluating our model. The first set if the evaluation of our trained models and the feature extraction procedures. These results are very important because it shows how feasible our current approach is and if we can rely on the prediction results. To do so we implement the three features described in sec.~\ref{sec:features} as well as the two classification models introduced in sec.~\ref{sec:classification-models}.

The second set of results describes different functional and non-functional tests we performed on the final stamp recognition model to make sure the system works correctly.

\begin{figure}
	\centering
	\includegraphics[width=0.5\textwidth]{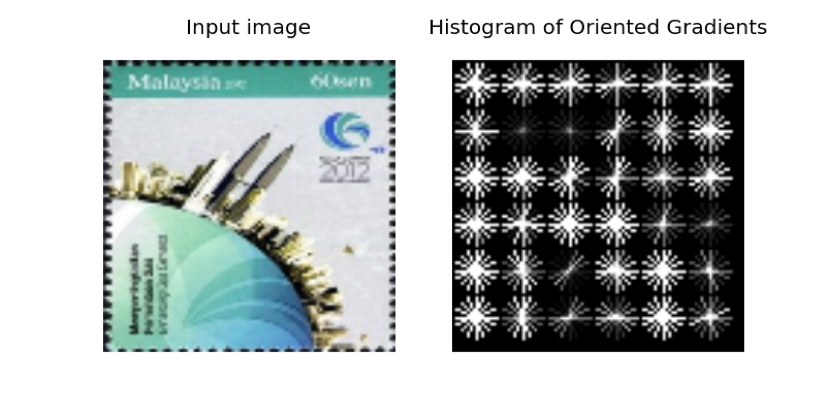}
	\includegraphics[width=0.5\textwidth]{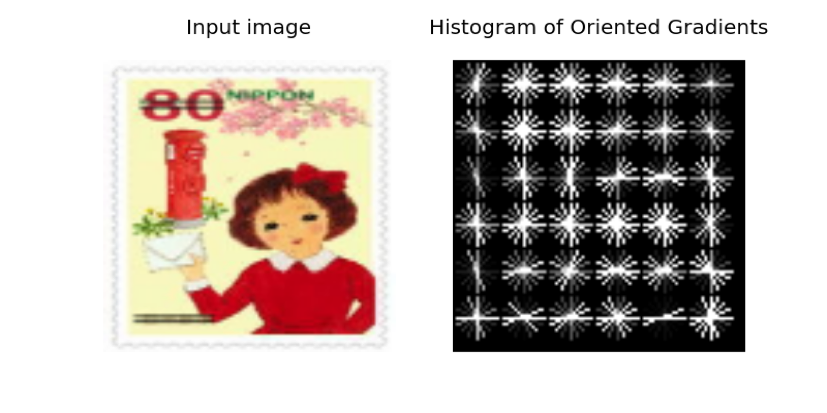}
	\newline
	\caption{Shows the example images with their corresponding HOG image}
		\label{fig:fig10}
\end{figure}

We first provide some visualized results of the extracted features. Fig.~\ref{fig:fig10} shows the result of applying HOG for feature extraction. It is interesting to see how well the HOG features have captured the texture.

\begin{figure}
	\centering
	\includegraphics[width=0.5\textwidth]{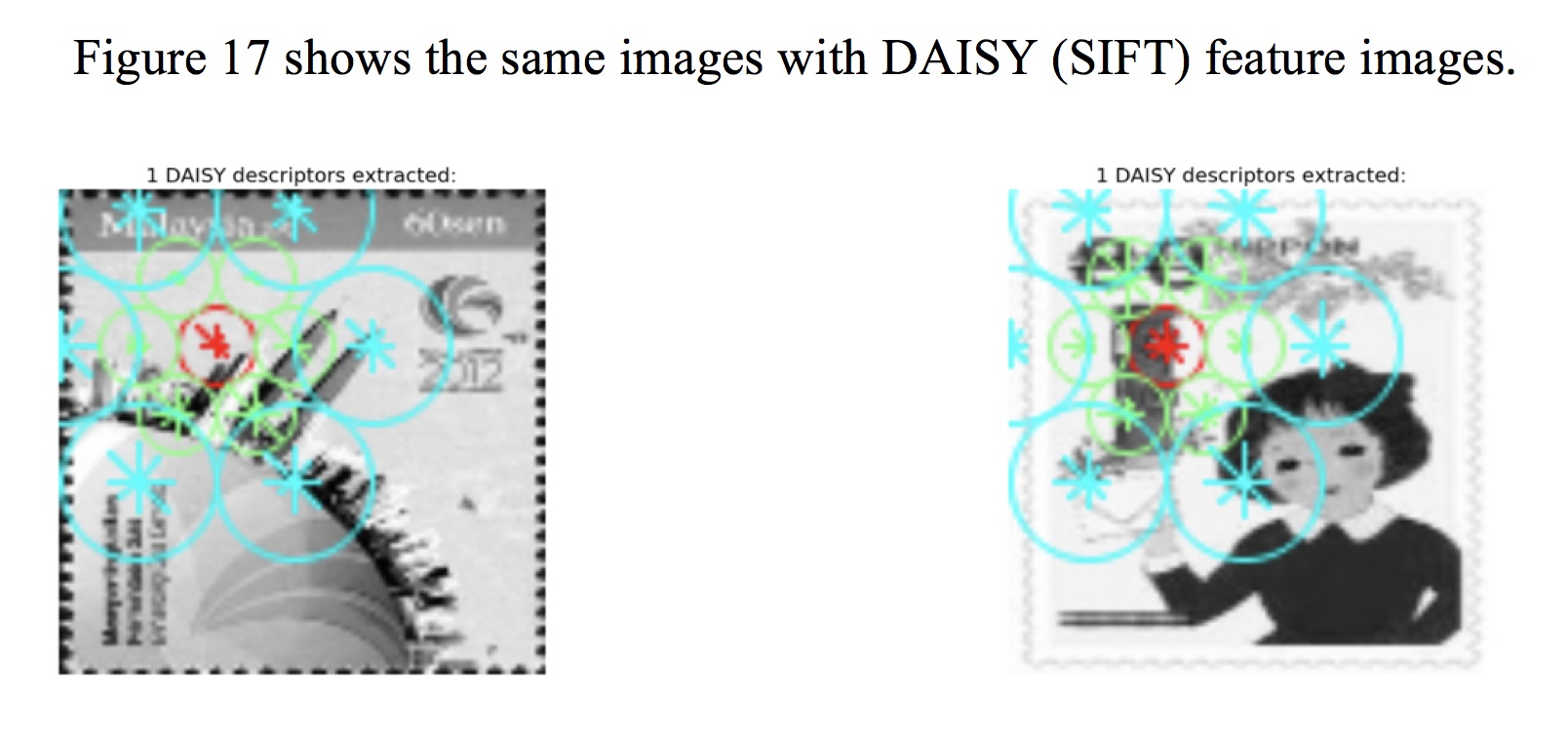}
	\newline
	\caption{Shows the example images with their corresponding HOG image}
	\label{fig:fig11}
\end{figure}

Fig.~\ref{fig:fig11} shows the same images with DAISY (SIFT) feature images.

\subsubsection{Evaluation result for country classification}

We implemented the logistic regression and SVM classifiers. For training we use the standard train-test split. We use 2/3 of the data from each category in training and 1/3 for testing. Note that the category is either the year or the country. Currently we have only implemented the classification with stand-alone features. Tables \ref{tab:table1}--\ref{tab:table6} show that confusion matrix for country classification with logistic regression and table shows the classification result with SVM. Since we randomly split data, we run the training and testing for 5 epochs, to make sure we are not biased to a certain train/test split.

\begin{center}
\begin{table}[h]
	\centering
	\resizebox{\textwidth}{!}
	{
		\centering
		\begin{minipage}{\textwidth}
			\begin{tabular}{|c|c|c|c|c|c|}
				\hline
				&\rot{China} &\rot{Japan}&\rot{Malaysia}&\rot{South-Korea}&\rot{Singapore}\\\hline		
				China&52&7&18&4&3\\\hline
				Japan&30&44&28&0&7\\\hline
				Malaysia&8&12&119&2&11\\\hline
				South-Korea&21&5&21&10&4\\\hline
				Singapore&1&5&60&1&34\\\hline
			\end{tabular}
		\end{minipage}
	}
	\newline
	\caption{Classification result for countries using Logistic Regression + Color Histogram for country model} 
	\label{tab:table1}
\end{table}
\end{center}

\begin{table}[h]
	\centering
	\resizebox{\textwidth}{!}
	{
		\centering
		\begin{minipage}{\textwidth}
			\begin{tabular}{|c|c|c|c|c|c|}\hline
				&\rot{China} &\rot{Japan}&\rot{Malaysia}&\rot{South-Korea}&\rot{Singapore}\\\hline		China&49&11&12&9&3\\\hline
				Japan&29&49&20&2&9\\\hline
				Malaysia&9&14&107&4&18\\\hline
				South-Korea&15&5&19&19&3\\\hline
				Singapore&2&7&30&1&61\\\hline
			\end{tabular}
			
		\end{minipage} 
	}
	\caption{Classification result for countries using SVM + Color Histogram for country model} 
	\label{tab:table2}
\end{table}
\begin{table}[h]
	\centering
	\resizebox{\textwidth}{!}
	{
		\centering
		\begin{minipage}{\textwidth}
			\begin{tabular}{|c|c|c|c|c|c|}\hline
				&\rot{China} &\rot{Japan}&\rot{Malaysia}&\rot{South-Korea}&\rot{Singapore}\\\hline		China&63&15&15&5&2\\\hline
				Japan&18&48&19&1&9\\\hline
				Malaysia&7&21&109&3&16\\\hline
				South-Korea&17&9&25&13&2\\\hline
				Singapore&0&7&48&1&34\\\hline
			\end{tabular}
			
		\end{minipage} 
	}
	\newline
	\caption{Classification result for countries using Logistic Regression + HOG for country model}
	\label{tab:table3} 
\end{table}
\begin{table}[h]
	\centering
	\resizebox{\textwidth}{!}
	{
		\centering
		\begin{minipage}{\textwidth}
			\begin{tabular}{|c|c|c|c|c|c|}\hline
				&\rot{China} &\rot{Japan}&\rot{Malaysia}&\rot{South-Korea}&\rot{Singapore}\\\hline		China&60&18&10&7&5\\\hline
				Japan&16&56&14&1&8\\\hline
				Malaysia&8&21&101&3&23\\\hline
				South-Korea&14&8&24&18&2\\\hline
				Singapore&0&6&32&1&51\\\hline
			\end{tabular}

		\end{minipage} 
	}
	\newline
	\caption{Classification result for countries using SVM + HOG for country model} 
	\label{tab:table4}
\end{table}
\begin{table}[h]
	\centering
	\resizebox{\textwidth}{!}
	{
		\centering
		\begin{minipage}{\textwidth}
			\begin{tabular}{|c|c|c|c|c|c|}\hline
				&\rot{China} &\rot{Japan}&\rot{Malaysia}&\rot{South-Korea}&\rot{Singapore}\\\hline		China&59&15&12&6&6\\\hline
				Japan&19&41&23&2&6\\\hline
				Malaysia&8&23&102&9&19\\\hline
				South-Korea&11&4&17&20&5\\\hline
				Singapore&1&20&45&2&31\\\hline
			\end{tabular}

		\end{minipage} 
	}
	\newline
	\caption{Classification result for countries using Logistic Regression + DAISY for country model} 
	\label{tab:table5}
\end{table}
\begin{table}[h]
	\centering
	\resizebox{\textwidth}{!}
	{
		\centering
		\begin{minipage}{\textwidth}
			\begin{tabular}{|c|c|c|c|c|c|}\hline
				&\rot{China} &\rot{Japan}&\rot{Malaysia}&\rot{South-Korea}&\rot{Singapore}\\\hline		China&57&17&12&8&5\\\hline
				Japan&15&48&17&3&8\\\hline
				Malaysia&10&15&106&8&22\\\hline
				South-Korea&12&5&16&22&2\\\hline
				Singapore&1&25&28&2&43\\\hline
			\end{tabular}

		\end{minipage} 
	}
	\newline
	\caption{Classification result for countries using SVM + DAISY for country model} 
	\label{tab:table6}
\end{table}

Based on the above we can see that the SVM classifier outperforms the logistic regression. This is shown in Table \ref{tab:table7} This is aligned with previous findings in machine learning literature which shows SVM is a better discriminative classifier among classical machine learning techniques.

\begin{table}[h]
	\centering
	\resizebox{\textwidth}{!}
	{
		\centering
		\begin{minipage}{\textwidth}
			\begin{tabular}{|c|c|c|c|c|c|}\hline
				{Classifier} &{Feature}&{Accuracy}\\\hline	
				Logistic Regression&1Color Histogram&51.0\\\hline
				Logistic Regression&HOG&52.6\\\hline
				Logistic Regression&DAISY&49.9\\\hline
				SVM&Color Histogram&56.2\\\hline
				SVM&HOG&6.4\\\hline
				SVM&DAISY&54.4\\\hline
			\end{tabular}

		\end{minipage} 
	}
	\newline
	\caption{Classification accuracy for different combination of features and classifiers for country model} 
	\label{tab:table7}
\end{table}

\subsubsection{Evaluation result for year classification}

Tables\ref{tab:table8}--\ref{tab:table13} show the performance of our approach considering different variations of features and classifiers. It is important to evaluate our trained models for different variations so we can choose the best model for the year classification model.

\begin{table}[h]
	\centering
	\resizebox{\textwidth}{!}
	{
		\centering
		\begin{minipage}{\textwidth}
			\begin{tabular}{|c|c|c|c|c|c|c|}\hline
				& 2010 & 2011 & 2012 & 2013 & 2014 & 2015\\\hline
				2010&20&14&15&12&10&12\\\hline
				2011&10&30&15&8&9&4\\\hline
				2012&12&12&30&6&10&11\\\hline
				2013&8&5&9&40&14&13\\\hline
				2014&4&10&8&12&24&14\\\hline
				2015&5&7&9&10&8&49\\\hline
			\end{tabular}

		\end{minipage} 
	}
			\newline
	    	\caption{Classification result for yeas using Logistic Regression + Color Histogram for year model} 
			\label{tab:table8}
\end{table}
\begin{table}[h]
	\centering
	\resizebox{\textwidth}{!}
	{
		\centering
		\begin{minipage}{\textwidth}
			\begin{tabular}{|c|c|c|c|c|c|c|}\hline
				& 2010 & 2011 & 2012 & 2013 & 2014 & 2015\\\hline
				2010&31&9&12&12&11&8\\\hline
				2011&8&38&8&12&10&4\\\hline
				2012&11&9&40&5&8&8\\\hline
				2013&5&6&6&66&2&4\\\hline
				2014&7&9&6&10&26&14\\\hline
				2015&6&5&10&8&6&53\\\hline
			\end{tabular}

		\end{minipage} 
	}
	\newline
	\caption{Classification result for countries using SVM + Color Histogram for year model} 
	\label{tab:table9}
\end{table}
\begin{table}[h]
	\centering
	\resizebox{\textwidth}{!}
	{
		\centering
		\begin{minipage}{\textwidth}
			\begin{tabular}{|c|c|c|c|c|c|c|}\hline
				& 2010 & 2011 & 2012 & 2013 & 2014 & 2015\\\hline
				2010&30&8&10&14&7&6\\\hline
				2011&8&38&8&12&10&14\\\hline
				2012&3&8&43&10&12&5\\\hline
				2013&9&5&8&59&1&7\\\hline
				2014&6&8&4&9&32&12\\\hline
				2015&8&5&6&10&9&50\\\hline
			\end{tabular}

		\end{minipage} 
	}
	\newline
	\caption{Classification result for countries using Logistic Regression + HOG for year model} 
	\label{tab:table10}
\end{table}
\begin{table}[h]
	\centering
	\resizebox{\textwidth}{!}
	{
		\centering
		\begin{minipage}{\textwidth}
			\begin{tabular}{|c|c|c|c|c|c|c|}\hline
				& 2010 & 2011 & 2012 & 2013 & 2014 & 2015\\\hline
				2010&37&9&10&14&7&6\\\hline
				2011&8&38&7&12&9&7\\\hline
				2012&5&6&50&10&6&4\\\hline
				2013&3&4&5&69&6&2\\\hline
				2014&3&6&5&10&40&8\\\hline
				2015&2&5&7&13&6&55\\\hline
			\end{tabular}

		\end{minipage} 
	}
	\newline
	\caption{Classification result for countries using SVM +
		 HOG for year model} 
	\label{tab:table11}
\end{table}
\begin{table}[h]
	\centering
	\resizebox{\textwidth}{!}
	{
		\centering
		\begin{minipage}{\textwidth}
			\begin{tabular}{|c|c|c|c|c|c|c|}\hline
				& 2010 & 2011 & 2012 & 2013 & 2014 & 2015\\\hline
				2010&14&16&15&14&9&5\\\hline
				2011&13&33&17&5&5&3\\\hline
				2012&8&16&27&12&6&12\\\hline
				2013&24&13&13&34&12&6\\\hline
				2014&2&9&7&14&19&21\\\hline
				2015&6&6&9&7&16&44\\\hline
			\end{tabular}

		\end{minipage} 
	}
	\newline
	\caption{Classification result for countries using Logistic Regression + DAISY for year model} 
	\label{tab:table12}
\end{table}
\begin{table}[h]
	\centering
	\resizebox{\textwidth}{!}
	{
		\centering
		\begin{minipage}{\textwidth}
			\begin{tabular}{|c|c|c|c|c|c|c|}\hline
				& 2010 & 2011 & 2012 & 2013 & 2014 & 2015\\\hline
				2010&347&10&9&17&9&4\\\hline
				2011&9&38&9&11&5&3\\\hline
				2012&5&6&42&13&5&10\\\hline
				2013&10&6&8&66&3&9\\\hline
				2014&2&4&6&14&37&9\\\hline
				2015&4&6&7&10&5&56\\\hline
			\end{tabular}

		\end{minipage} 
	}
	\newline
	\caption[Table 13]{Classification result for countries using Logistic Regression + DAISY for year model}
	\label{tab:table13}
\end{table}

Based on the above we can see that the SVM classifier outperforms the logistic regression. This is shown in Table  \ref{tab:table14}. This is aligned with previous findings in machine learning literature which shows SVM is a better discriminative classifier among classical machine learning techniques. 

\begin{table}[h]
	\centering
	\resizebox{\textwidth}{!}
	{
		\centering
		\begin{minipage}{\textwidth}
			\begin{tabular}{|c|c|c|c|}\hline
				{Classifier} &{Feature}&{Accuracy}\\\hline	
				Logistic Regression&1Color Histogram&37.5\\\hline
				Logistic Regression&HOG&40.1\\\hline
				Logistic Regression&DAISY&35.6\\\hline
				SVM&Color Histogram&44.9\\\hline
				SVM&HOG&45.3\\\hline
				SVM&DAISY&42.3\\\hline
			\end{tabular}

		\end{minipage} 
	}
	\newline
	\caption[Table 14]{Classification accuracy for different combination of features and classifiers for year model} 
	\label{tab:table14}
\end{table}

\subsubsection{Efficiency of the stamp recognition system}

One important property of a well-designed system is the efficiency. In this section we present the efficiency of the system in terms of the processing time for different functionalities of the system. This is shown in Table 15.

\subsubsection{Robustness of the classification model}

Another important property of a machine learning model is how robust is the model with respect to small perturbation of the images. One standard way to make the models robust to these perturbations is to augment the training dataset with rotated and flipped images. This is well studied in (Brownlee), (A), and (Wei). We uses the similar strategy and augment the dataset by adding flipped images and rotated images with 90 and 180 degree rotations to the dataset. Evaluations showed that with this technique we are able to recognize images if they have a slight rotation.

\section*{Conclusion}

In this report we present our research on the stamp recognition problem. Stamp recognition is a fine-grained classification problem in computer vision which is categorized as a supervised learning problem.
We proposed a classification model which uses a classical machine learning approach to classify stamps to five different countries and 5 different years. To do that we used three modern feature extraction techniques, 1) Color histogram, 2) Histogram of Oriented Gradients, 3) DAISY features. These three features capture both color and texture features of an image and highly recommended in computer vision research. 
We used two famous classification models 1) Logistic Regression, and 2) Support Vector Machines (SVM). Both of these models are well-studied in the literature. 
Our evaluation shows that considering the challenging problem of stamp recognition, we can achieve a reasonable accuracy for classification of countries and years. Our best result achieved using SVM and all three features we obtained from the image. 
Despite our good results, we believe using modern deep learning techniques which extract richer features from the image helps improving the results.

{\small
	\bibliographystyle{ieee}
	\bibliography{library}

}

\end{document}